\documentclass[11pt,a4paper]{article}
\usepackage[hyperref]{emnlp2020}
\usepackage{times}
\usepackage{latexsym}

\usepackage{booktabs}
\usepackage{cleveref}
\usepackage{subfigure}
\usepackage{graphicx}

\usepackage{hyperref}

\usepackage{paralist}
\usepackage{tablefootnote}

\usepackage{microtype}

\aclfinalcopy 

\title{Text Segmentation by Cross Segment Attention}

\author{
    Michal Lukasik, 
    Boris Dadachev,
    Gon\c{c}alo Sim\~{o}es, 
    Kishore Papineni\\
  Google Research \\
  \texttt{\{mlukasik,bdadachev,gsimoes,papineni\}@google.com}}

\date{}

\begin{document}
\maketitle
\begin{abstract}
Document and discourse segmentation are two fundamental NLP tasks pertaining to breaking up text into constituents,
which are commonly used to help downstream tasks such as information retrieval or text summarization.
In this work, we propose three transformer-based architectures and provide comprehensive comparisons with previously proposed approaches on three standard datasets. We establish a new state-of-the-art, reducing in particular the error rates by a large margin in all cases. We further analyze model sizes and find that we can build models with many fewer parameters while keeping good performance, thus facilitating real-world applications.
\end{abstract}

\section{Introduction}
Text segmentation is a traditional NLP task that breaks up text into constituents, according to predefined requirements.
It can be applied to documents, in which case the objective is to create logically coherent sub-document units. 
These units, or segments, can be any structure of interest, such as paragraphs or sections. This task is often referred to as \emph{document segmentation} or sometimes simply \emph{text segmentation}. 
In \Cref{fig:text_seg_example} we show one example of document segmentation from Wikipedia, on which the task is typically evaluated \cite{koshorek18,badjatiya2018}.

Documents are often multi-modal, in that they cover multiple aspects and topics; breaking a document into uni-modal segments can help improve and/or speed up down stream applications.
For example, document segmentation has been shown to improve information retrieval by indexing sub-document units instead of full documents \cite{Llopis2002, Shtekh2018}.
Other applications such as summarization and information extraction can also benefit from text segmentation \cite{koshorek18}.

\begin{figure}[t!]
  \center{\footnotesize
  \setlength{\tabcolsep}{0.0in}
    \begin{tabular}{p{7.9cm}}
    \hline 
      
    \textbf{Early life and marriage}: 
    
    Franklin Delano Roosevelt was born on January 30, 1882, in the Hudson Valley town of Hyde Park, New York, to businessman James Roosevelt I and his second wife, Sara Ann Delano. (...)
    
    Aides began to refer to her at the time as ``the president's girlfriend'', and gossip linking the two romantically appeared in the newspapers.
    
    \centerline{(...)}
    
    \textbf{Legacy}: 
    
    Roosevelt is widely considered to be one of the most important figures in the history of the United States, as well as one of the most influential figures of the 20th century. (...) Roosevelt has also appeared on several U.S. Postage stamps.
    
    \\ \hline
    \end{tabular}     
  }
  \vspace{-0.1in}
  \caption{ Illustration of text segmentation on the example of the Wikipedia page of President Roosevelt.  The aim of document segmentation is breaking the raw text into a sequence of logically coherent sections (e.g., ``Early life and marriage'' and ``Legacy'' in our example).}
  \vspace{-0.2cm}
  \label{fig:text_seg_example}
\end{figure}

A related task called \emph{discourse segmentation} breaks up pieces of text into sub-sentence elements called Elementary Discourse Units (\emph{EDUs}). EDUs are the minimal units in discourse analysis according to the Rhetorical Structure Theory \cite{mann1988}. 
In \Cref{fig:discourse_seg_example} we show examples of EDU segmentations of sentences.
For example, the sentence ``Annuities are rarely a good idea at the age 35 because of withdrawal restrictions'' decomposes into the following two EDUs: ``Annuities are rarely a good idea at the age 35'' and ``because of withdrawal restrictions'', the first one being a statement and the second one being a justification in the discourse analysis.
In addition to being a key step in discourse analysis \cite{joty-etal-2019-discourse}, discourse segmentation has been shown to improve a number of downstream tasks, such as text summarization, by helping to identify fine-grained sub-sentence units that may have different levels of importance when creating a summary \cite{li-etal-2016-role}.

\begin{figure}[t!]
  \center{\footnotesize
  \setlength{\tabcolsep}{0.0in}
    \begin{tabular}{p{7.9cm}}
    \hline 
      
    \textbf{Sentence 1}: 
    
    %
    Annuities are rarely a good idea at the age 35 $\Vert$ because of withdrawal restrictions
    \\ \hline
    \textbf{Sentence 2}: 
    
    %
    Wanted: $\Vert$ An investment $\Vert$ that's as simple and secure as a certificate of deposit $\Vert$ but offers a return $\Vert$ worth getting excited about.
    \\ \hline
    \end{tabular}     
  }
  \vspace{-0.1in}
  \caption{Example discourse segmentations from the RST-DT dataset~\cite{carlson:sigdial2001}. In the segmentations, the EDUs are separated by the $\Vert$ character.}
  \vspace{-0.2cm}
  \label{fig:discourse_seg_example}
\end{figure}

Multiple neural approaches have been recently proposed for document and discourse segmentation.
\citet{koshorek18} proposed the use of hierarchical Bi-LSTMs for document segmentation.
Simultaneously, \citet{segbot} introduced an attention-based model for both document segmentation and discourse segmentation, and \citet{wang:emnlp18} obtained state of the art results on discourse segmentation using pretrained contextual embeddings \citep{elmopaper}. Also, a new large-scale dataset for document segmentation based on Wikipedia was introduced by \citet{koshorek18}, providing a much more realistic setup for evaluation than the previously used small scale and often synthetic datasets such as the Choi dataset \cite{Choi2000}.

However, these approaches are evaluated on different datasets and as such have not been compared against one another.
Furthermore they mostly rely on RNNs instead of the more recent transformers \cite{Attention2017} and in most cases do not make use of contextual embeddings which have been shown to help in many classical NLP tasks \cite{BertPaper}. 

In this work we aim at addressing these limitations and bring the following contributions:
\begin{compactenum}
\item We compare recent approaches that were proposed independently for text and/or discourse segmentation \cite{segbot,koshorek18,wang:emnlp18} on three public datasets.
\item We introduce three new model architectures based on transformers and BERT-style contextual embeddings to the document and discourse segmentation tasks. We analyze the strengths and weaknesses of each architecture and establish a new state-of-the-art.
\item We show that a simple paradigm argued for by some of the earliest text segmentation algorithms can achieve competitive performance in the current neural era.
\item We conduct ablation studies analyzing the importance of context size and model size.
\end{compactenum}

\section{Literature review}

\paragraph{Document segmentation}
Many early research efforts were focused on unsupervised text
segmentation, doing so by quantifying lexical cohesion within small
text segments \cite{hearst97,Choi2000}. 
Being hard to precisely define and quantify, lexical cohesion has often
been approximated by counting word repetitions.
Although computationally expensive, unsupervised Bayesian approaches have also been popular 
\cite{utiyama01,eisenstein09,mota19}.
However, unsupervised algorithms suffer from two main drawbacks: they
are hard to specialize for a given domain and in most cases do not naturally deal
with multi-scale issues. Indeed, the desired segmentation granularity 
(paragraph, section, chapter, etc.) is necessarily task dependent and supervised learning provides
a way of addressing this property.
Therefore, supervised algorithms have been a focus of many recent works. 

In particular, multiple neural approaches have been proposed for the task. 
In one, a sequence labeling algorithm is proposed where each sentence is encoded using a Bi-LSTM over tokens, and then a Bi-LSTM over sentence encodings is used to label each sentence as ending a segment or not \cite{koshorek18}. 
Authors consider a large dataset based on Wikipedia, and report improvements over unsupervised text segmentation methods.
In another work, a sequence-to-sequence model is proposed \cite{segbot}, where the input is encoded using a BiGRU and segment endings are generated using a pointer network \cite{Vinyals2015}. The authors report significant improvements over sequence labeling approaches, however on a dataset composed of 700 artificial documents created by concatenating segments from random articles from the Brown corpus \cite{Choi2000}.
Lastly, \citet{badjatiya2018} consider an attention-based CNN-Bi-LSTM model and evaluate it on three small-scale datasets.


\paragraph{Discourse Segmentation}

Contrary to document segmentation, discourse segmentation has historically been framed as a supervised learning task. However, a challenge of applying supervised approaches for this type of segmentation is the fact that the available dataset for the task is limited \cite{carlson:sigdial2001}. 
For this reason, approaches for discourse segmentation usually rely on external annotations and resources to help the models generalize.
Early approaches to discourse segmentation were based on features from linguistic annotations such as POS tags and parsing trees \cite{soricut-marcu-2003-sentence,xuan-bach-etal-2012-reranking,joty-etal-2015-codra}. The performance of these systems was highly dependent on the quality of the annotations.

Recent approaches started to rely on end-to-end neural network models that do not need linguistic annotations to obtain high-quality results, relying instead on pretrained models to obtain word or sentence representations. 
An example of such work is by \citet{segbot}, which proposes a sequence-to-sequence model getting a sequence of GloVe \cite{pennington-etal-2014-glove} word embeddings as input and generating the EDU breaks. 
Another approach utilizes ELMO pretrained embeddings in the CRF-Bi-LSTM architecture and achieves state-of-the-art results on the task \cite{wang:emnlp18}.

\section{Architectures}
\label{sec:model_arch}

We propose three model architectures for segmentation. One uses only local context around each candidate break, while the other two leverage the full context from the input (by candidate break, we mean any potential segment boundary).  

All our models rely on the same preprocessing technique and simply feed the raw input into a word-piece (sub-word) tokenizer \cite{wu16}.
We use the word-piece tokenizer implementation that was open-sourced as part of the BERT release \cite{BertPaper}, 
more precisely its English, uncased variant, which has a vocabulary size 
of 30,522 word-pieces.

\begin{figure*}
\center
\subfigure[Cross-Segment BERT]{\includegraphics[width=0.35\textwidth]{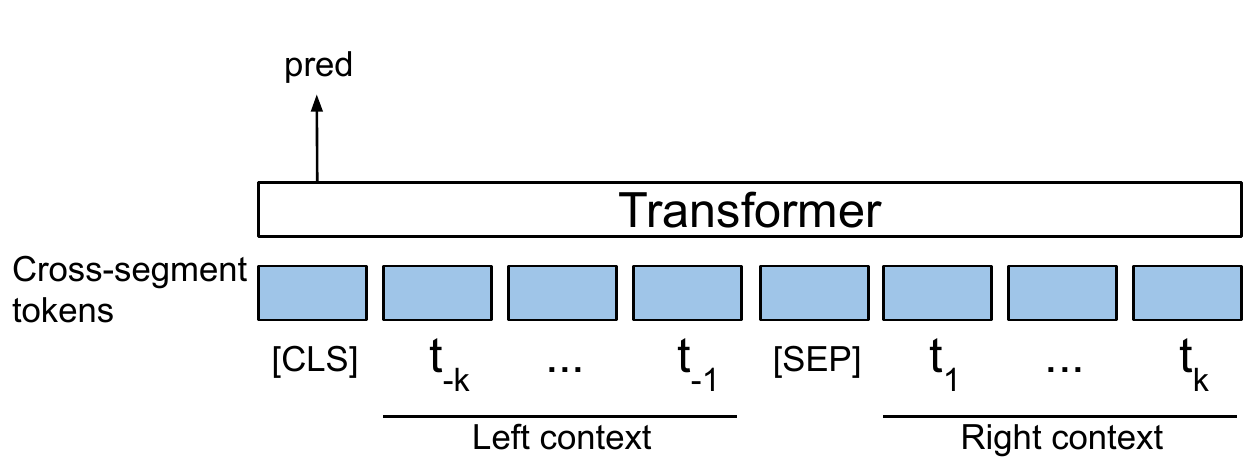}\label{fig:crossBert}}
    \quad
\subfigure[BERT+Bi-LSTM]{\includegraphics[width=0.24\textwidth]{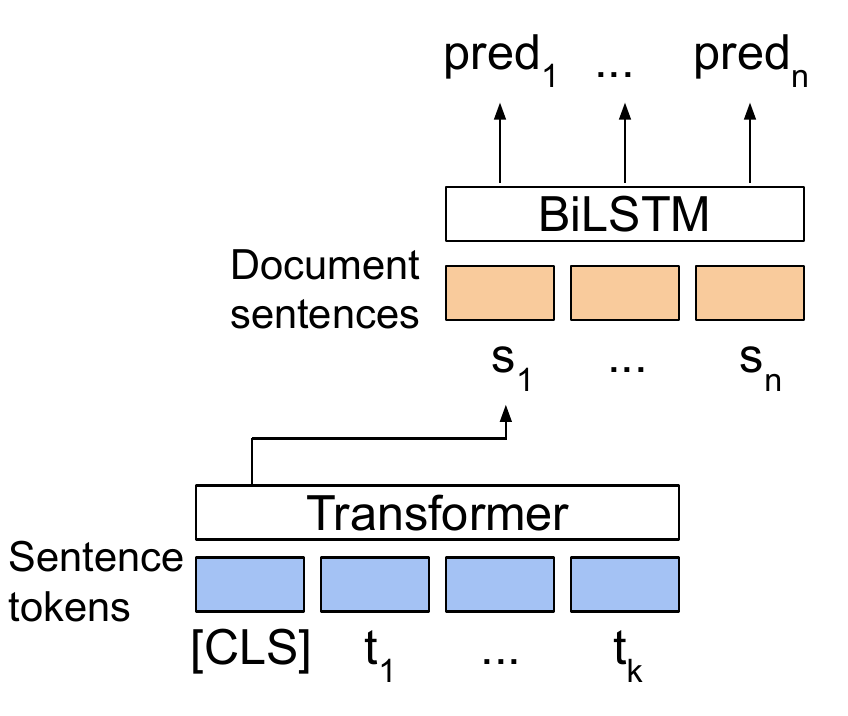}\label{fig:lstmBert}}
    \quad
\subfigure[Hierarchical BERT]{\includegraphics[width=0.24\textwidth]{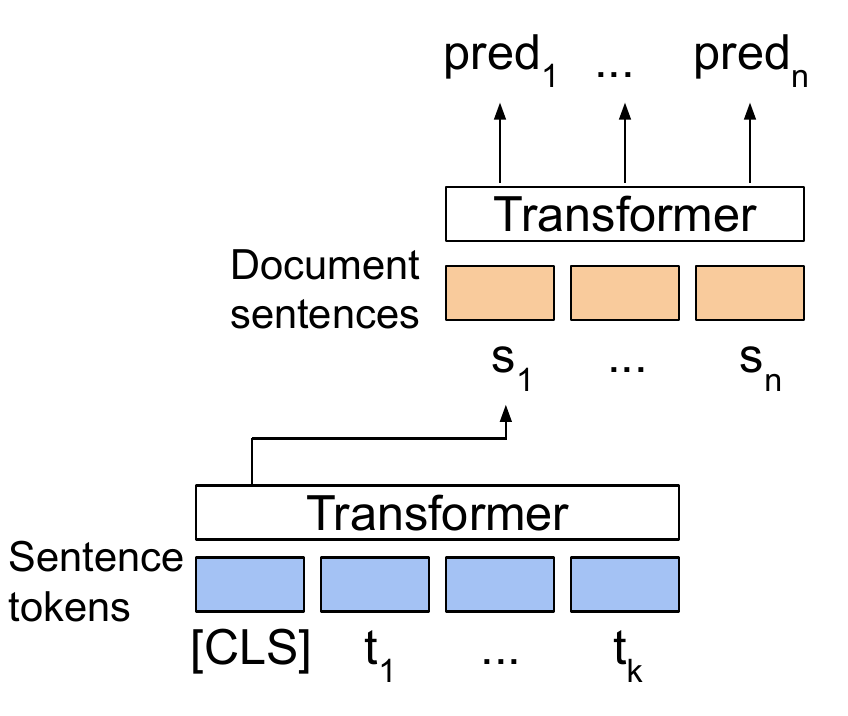}\label{fig:hiBert}}
    \caption{Our proposed segmentation models, illustrating the document segmentation task. In the cross-segment BERT model (left), we feed a model with a local context surrounding a potential segment break: k tokens to the left and k tokens to the right. In the BERT+Bi-LSTM model (center) we first encode each sentence using a BERT model, and then feed the sentence representations into a Bi-LSTM. In the hierarchical BERT model (right), we first encode each sentence using BERT and then feed the output sentence representations in another transformer-based model.}
\end{figure*}

\subsection{Cross-segment BERT}


For our first model, we represent each candidate break by its left and right local contexts, i.e., the sequences of word-piece tokens that come before and after, respectively, the candidate break.
The main motivation for this model is its simplicity; however, using only local contexts might be sub-optimal, as longer distance linguistic artifacts are likely to help locating breaks.
Using such a simple model is a departure from recent trends favoring hierarchical models, which are conceptually appealing to model documents.
However, it is also interesting to note that using local context was a common approach with earlier text segmentation models, such as \cite{hearst97}, which were studying semantic shift by comparing the word distributions before and after each candidate break.

In \Cref{fig:crossBert} we illustrate the model. 
The input is composed of a \emph{[CLS]} token, followed by the two contexts concatenated together, and separated by a \emph{[SEP]} token. When necessary, short contexts are padded to the left or to the right with \emph{[PAD]} tokens. \emph{[CLS]}, \emph{[SEP]} and \emph{[PAD]}
are special tokens introduced by BERT \cite{BertPaper}. They stand for, respectively, "classification token" (since it is typically for classification tasks, as a representation of the entire input sequence), "separator token" and "padding token".
The input is then fed into a transformer encoder \cite{Attention2017}, which is initialized
with the publicly available BERT$_\textrm{\sc Large}$ model. The BERT$_\textrm{\sc Large}$ model has 24 layers, uses
1024-dimensional embeddings and 16 attention heads.
The model is then fine-tuned on each task. 
The released BERT checkpoint supports sequences of up to 512 tokens, so we keep at most 255 word-pieces for each side. We study the effect of length of the contexts, and denote the context configuration by $n$-$m$ where $n$ and $m$ are the number of word piece tokens before and after the \emph{[SEP]} token. 

\subsection{BERT+Bi-LSTM}

Our second proposed model is illustrated in \Cref{fig:lstmBert}.
It starts by encoding each sentence with BERT$_\textrm{\sc Large}$ independently. Then, the tensors produced for each sentence are fed into a Bi-LSTM that is responsible for capturing a representation of the sequence of sentences with an indefinite size.

When encoding each sentence with BERT, all the sequences start with a \emph{[CLS]} token. 
If the segmentation decision is made at the sentence level (e.g., document segmentation), we use the \emph{[CLS]} token as input of the LSTM. 
In cases in which the segmentation decision is made at the word level (e.g., discourse segmentation), we obtain BERT's full sequence output and use the left-most word-piece of
each word as an input to LSTM.
Note that, due to the context being short for the discourse segmentation task,
it is fully encoded in a single pass using BERT. Alternatively, one could encode
each word independently; considering that many words consist of a single
word-piece, encoding them with a deep transformer
encoder would be somewhat wasteful of computing resources.

With this model, we reduce the BERT's inputs to a maximum sentence size of 64 tokens.
Keeping this size small helps reduce training and inference times, since the computational 
cost of transformers (and self-attention in particular) increases quadratically with the input length.
Then, the LSTM is responsible for handling the diverse and potentially large sequence of sentences with linear computational complexity. 
In practice, we set a maximum document length of 128 sentences.
Longer documents are split into consecutive, non-overlapping chunks of 128 sentences and treated as independent documents.

In essense, the hierarchical nature of this model is close to the recent neural approaches such
as \cite{koshorek18}.

\subsection{Hierarchical BERT}

Our third model is a hierarchical BERT model
that also encodes full documents, replacing
the document-level LSTM encoder from the BERT+Bi-LSTM model with a
transformer encoder. This architecture is similar to the HIBERT 
model used for document summarization \cite{zhang19}, encoding each sentence independently.
The $[CLS]$ token representations from sentences are passed into the document encoder,
which is then able to relate the different sentences through cross-attention, as
illustrated in \Cref{fig:hiBert}.

Due to the quadratic computational cost of transformers, we use the same limits as BERT+Bi-LSTM for input sequence sizes: 64 word-pieces per sentence and 128 sentences per document.

To keep the number of model parameters comparable with our other
proposed models, we use 12 layers for both the sentence and the
document encoders, for a total of 24 layers. In order to use the BERT$_\textrm{Base}$ checkpoint for these experiments, we use 12 attention heads and 768-dimensional word-piece embeddings.

We study two alternative initialization procedures:
\begin{compactitem}
    \item initializing both sentence and document encoders using BERT$_\textrm{Base}$
    \item pre-training all model weights on Wikipedia, using the procedure described in \cite{zhang19}, which can be summarized as a "masked sentence" prediction objective, analogously to the
    "masked token" pre-training objective from BERT.
\end{compactitem}
We call this model hierarchical BERT for consistency with the literature.

\section{Evaluation methodology}

\subsection{Datasets}
\label{ssec:datasets}

We perform our experiments on datasets commonly used in the literature.
Document segmentation experiments are done on Wiki-727K and Choi,
while discourse segmentation experiments are done on the RST-DT dataset.
We summarize statistics about the datasets in \Cref{tab:data}.


\paragraph{Wiki-727K}
The Wiki-727K dataset \cite{koshorek18} contains 727 thousand articles from a snapshot 
of the English Wikipedia, which are randomly partitioned into train, development and test sets.
We re-use the original splits provided by the authors. 
While several segmentation
granularities are possible, the dataset is used to predict \textit{section} boundaries.
The average number of segments per document is $3.5$, with an average segment length of $13.6$ sentences.

We found that the preprocessing methodology used on the Wiki-727K dataset can have a noticeable effect on the final numerical results, in particular when filtering lists, code snippets and other special elements. We used the original preprocessing script \cite{koshorek18} for a fair comparison.

\paragraph{Choi}

Choi's dataset \cite{Choi2000} is an early dataset containing 700 synthetic documents made of concatenated extracts of news articles.
Each document is made of 10 segments, where each segment was created by sampling a document from the Brown corpus
and then sampling a random segment length up to 11 sentences.

This dataset was originally used to evaluate unsupervised segmentation algorithms, so it is somewhat ill-designed to evaluate supervised algorithms.
We use this dataset as a best-effort attempt to allow comparison with some of the previous literature. However, we had to create our own splits as no standard splits exist: we randomly sampled 200 documents as a test set and 50 documents as a validation set, leaving 450 documents for training, following evaluation from \citet{segbot}. 
Since the Brown corpus only contains 500 documents, the same documents are sampled over and over, necessarily resulting in data leakage between the different splits. Its use should therefore be discouraged in future research.


\paragraph{RST-DT}
We perform experiments on discourse segmentation on the RST Discourse Treebank (RST-DT) \cite{carlson:sigdial2001}. The dataset is composed of 385 Wall Street Journal articles that are part of the Penn Treebank \cite{Marcus:1994:PTA:1075812.1075835}, and is split into the train set composed of 347 articles and the test set composed of 38 articles. 
We found that the choice of a validation set (held out from the train set) has a large
impact on model performance. For this reason, we conduct 10-fold cross validation and report the average over test set metrics.

Since this dataset is used for discourse segmentation, all the segmentation decisions are made at the intra-sentence level (i.e., the context that is used in the decisions is just a sentence). In order to make the evaluation consistent with other systems from the literature we decided to use the sentence splits that are available in the dataset, even though they are not human annotate. For this reason, there are cases in which some EDUs (which were manually annotated) overlap between two sentences. In such cases, we merge the two sentences.

\subsection{Metrics}

Following the trend of many studies on text segmentation \cite{soricut-marcu-2003-sentence, segbot}, we evaluate our approaches using Precision, Recall and F1-score with regard to the internal boundaries of the segments only. In our evaluation we do not include the last boundary of each sentence/document, because it would be trivial to categorize it as a positive boundary, which would lead to an artificial inflation of the results.

To allow comparison with the existing literature, we also use the $P_k$ metric \cite{beeferman99} 
to evaluate our results on the Choi's dataset (note that lower $P_k$ scores indicate better performance). 
$k$ is set, as is customary, to half the average segment size over the reference segmentation. 
The $P_k$ metric is less harsh than the F1-score in that it takes into account near misses. It is important to note that $P_k$ metric is known to suffer from biases, for example penalizing false negatives more than false positives and discounting errors close to the document extremities \cite{pevzner02}.


\begin{table}
\footnotesize
\begin{center}
\begin{tabular}{lccc}
\toprule 
 & Docs & Sections & Sentences \\
\midrule 
Wiki-727K Train & 582,146 & 2,025,358 & 26,988,063\\
Wiki-727K Dev & 72,354 & 179,676 & 3,375,081\\
Wiki-727K Test & 73,233 & 182,563 & 3,457,771\\
\midrule
Choi Train & 450 & 4,500 & 31,075\\
Choi Dev & 50 & 500 & 3,291\\
Choi Test & 200 & 2,000 & 14,039\\
\midrule
 & Docs & Sentences & EDUs \\
\midrule
RST-DT Train & 347 & 7,028 & 19,443 \\
RST-DT Test & 38 & 864 & 2,346 \\
\bottomrule 
\end{tabular}
\end{center}
\caption{Statistics about the datasets.}
\label{tab:data}
\end{table}

\section{Results}

\begin{table*}
\footnotesize
\begin{center}
\tabcolsep=0.11cm
\begin{tabular}{l | ccc | ccc | cc}
\toprule 
 & \multicolumn{3}{c|}{Wiki-727K}  & \multicolumn{3}{c|}{RST-DT} & \multicolumn{2}{c}{Choi}  \\ 
 & \multicolumn{1}{c}{Precision} & \multicolumn{1}{c}{Recall} & \multicolumn{1}{c|}{F1} & \multicolumn{1}{c}{Precision} & \multicolumn{1}{c}{Recall} & \multicolumn{1}{c|}{F1} & \multicolumn{1}{c}{F1} & \multicolumn{1}{c}{P$_k$} \\ 
\midrule 
Bi-LSTM \cite{koshorek18} 
& 69.3$\pm$0.1 & 49.5$\pm$0.2 & 57.7$\pm$0.1 & - & - & - & - & - \\ 


SEGBOT \cite{segbot} 
& - & - & - & 91.6 & 92.8 & 92.2 & - & 0.33 \\ 

Bi-LSTM+CRF \cite{wang:emnlp18} 
& - & - & - & 92.8 & 95.7 & 94.3 & - & -\\

\midrule


Cross-segment BERT 128-128 & 69.1$\pm$0.1 & 63.2$\pm$0.2 & 66.0$\pm$0.1 & 92.1$\pm$0.8 & \bf98.0$\pm$0.4 &  95.0$\pm$0.5 & \bf 99.9$\pm$0.1 & \bf 0.07$\pm$0.04  \\ 
BERT+Bi-LSTM & 67.3$\pm$0.1 & 53.9$\pm$0.1 & 59.9$\pm$0.1 & \bf 94.4$\pm$0.5 & 96.0$\pm$0.4 & \bf95.2$\pm$0.3 & 99.8$\pm$0.1 & 0.17$\pm$0.06  \\
Hier. BERT & \bf69.8$\pm$0.1 & \bf63.5$\pm$0.1 & \bf66.5$\pm$0.1 & 93.8$\pm$0.7 & 96.7$\pm$0.5 & \bf95.2$\pm$0.4 & 99.5$\pm$0.1 & 0.38$\pm$0.09 \\
\midrule
Human \cite{wang:emnlp18} & - & - & - & 98.3 & 98.2 & 98.5 & - & - \\
\bottomrule 
\end{tabular}
\end{center}
\caption{Test set results on text segmentation and discourse segmentation for baselines and our models. Where possible, we estimate standard deviations by bootstrapping the test set 100 times.}
\label{tab:results-all}
\end{table*}

In \Cref{tab:results-all}, we report results from the document
and discourse segmentation experiments on the three datasets presented
in \Cref{ssec:datasets}.
We include several state-of-the-art baselines which had not been compared against one another before, as they have been proposed
independently over a short time period: hierarchical Bi-LSTM \cite{koshorek18}, SEGBOT \cite{segbot} and Bi-LSTM+CRF+ELMO \cite{wang:emnlp18}. We also include the human annotation baseline from \cite{wang:emnlp18}, providing an additional reference point on the RST-DT dataset to the trained models.
We estimate standard deviations for our proposed models
and were able to calculate them from the hierarchical Bi-LSTM, whose code and trained checkpoint were publicly released. 

To train our models, we used the AdamW optimizer \cite{adamw18} with 
a 10\% dropout rate as well as a linear warmup procedure. Learning rates are set 
between 1e-5 and 5e-6, chosen to maximize the F1-score on the validation sets 
from each dataset. For the more expensive models, and especially on the Wiki-727K dataset,
we trained our models using Google Cloud TPUs.

We can see from the table that our models outperform the baselines across all datasets, 
reducing the relative error margins from the best baseline by $20\%$, $16\%$ and $79\%$ 
respectively on the Wiki-727K, RST-DT and Choi datasets.
The improvements are statistically significant for all datasets. 
The errors are impressively low on the Choi dataset, but 
it is important to point out that it is a small-scale synthetic dataset, and as such limited. Since each document
is a concatenation of extracts from random news articles, it is an artificially easy task
for which a previous neural baseline achieved an already low error margin. 
Moreover, on this dataset, the cross-segment BERT model obtains very good results compared to the hierarchical models which do not attend across the candidate break.
This aligns with the expectation that locally attending across a segment break is sufficient here, as we expect large semantic shifts due to the artificial nature of the dataset.

Hierarchical models, with a sentence encoder followed by a document encoder, perform well 
on the RST-DT dataset. As a reminder, this discourse segmentation task is about segmenting individual sentences so there is no notion of document context.
In order to study whether the hierarchical structure is really necessary for discourse segmentation, we also trained a model without the Bi-LSTM (that is, making predictions directly using BERT): this decreased the F1-score by 0.4\%.
It is also worth noting that several known LSTM downsides were particularly apparent on the Wiki-727K dataset: the model was harder to train and significantly slower during both training and inference.

Regarding the hierarchical BERT model, different initialization methods were used for the two document segmentation datasets. 
On the Choi dataset, a HIBERT initialization (a model fully pre-trained end-to-end for hierarchical BERT, similarly to \cite{zhang19}
was necessary to get good results, due the small dataset size. 
On the contrary, we obtained slightly better results
initializing both levels of the hierarchy with BERT$_\textrm{Base}$ on the Wiki-727K dataset,
even though the model took longer to converge. Other initializations,
e.g., random for both levels of the hierarchy or BERT$_\textrm{Base}$ at the lower level and random at the upper level, gave worse results.


Perhaps the most surprising result from \Cref{tab:results-all} is the good performance of our
cross-segment BERT model across all datasets, since it only relies on local context to make predictions.
And while the BERT checkpoints were pre-trained using (among other things) the next-sentence prediction
task, it was not clear a priori that our cross-segment BERT model 
would be able to detect much more subtle semantic shifts.
To further evaluate the effectiveness of this model, we tried using longer contexts. 
In particular, we considered using a cross-segment BERT with 255-255 contexts, achieving $67.1$ F1, $73.9$ recall and $61.5$ precision scores.
Therefore, we can see that encoding the full document in a hierarchical manner using transformers 
does not improve over cross-segment BERT on this dataset. 
This suggests that BERT self-attention mechanism applied across candidate segment breaks, with a limited context, is in this case just as powerful as separately encoding each sentence and then allowing a flow of information across encoded sentences.
In the next section we further analyze the impact of context length on the results from the cross-segment BERT model.

\section{Analyses}
\label{sec:analyses}

In this section we perform additional analyses and ablation studies to better understand our segmentation models.

Experiments revolve around the cross-segment BERT model. We choose this model because it has several advantages over
its alternatives:
\begin{compactitem}
\item It outperforms all baselines previously reported as state-of-the-art, and its results are competitive with the more complex hierarchical approaches we considered.
\item It is conceptually close to the original BERT model \cite{BertPaper}, whose code is open-source,
and is as such simple to implement.
    \item It only uses local document context and therefore does not require encoding an entire document to segment
a potentially small piece of text of interest. 
\end{compactitem}

One application for text segmentation is in assisting a document writer in composing a document, for example to save them time 
and effort.
The task proposed by \citet{lukasik2018}, aligned with what industrial applications such as Google Docs Explore provide, was to recommend related entities to a writer in real time.
However, text segmentation could also help authors in structuring their document better by suggesting where a section break might be appropriate. 
Motivated by this application, we next analyze how much context is needed to reliably predict a section break.

\subsection{Role of trailing context size}

\begin{figure}
\center
\includegraphics[width=0.5\textwidth]{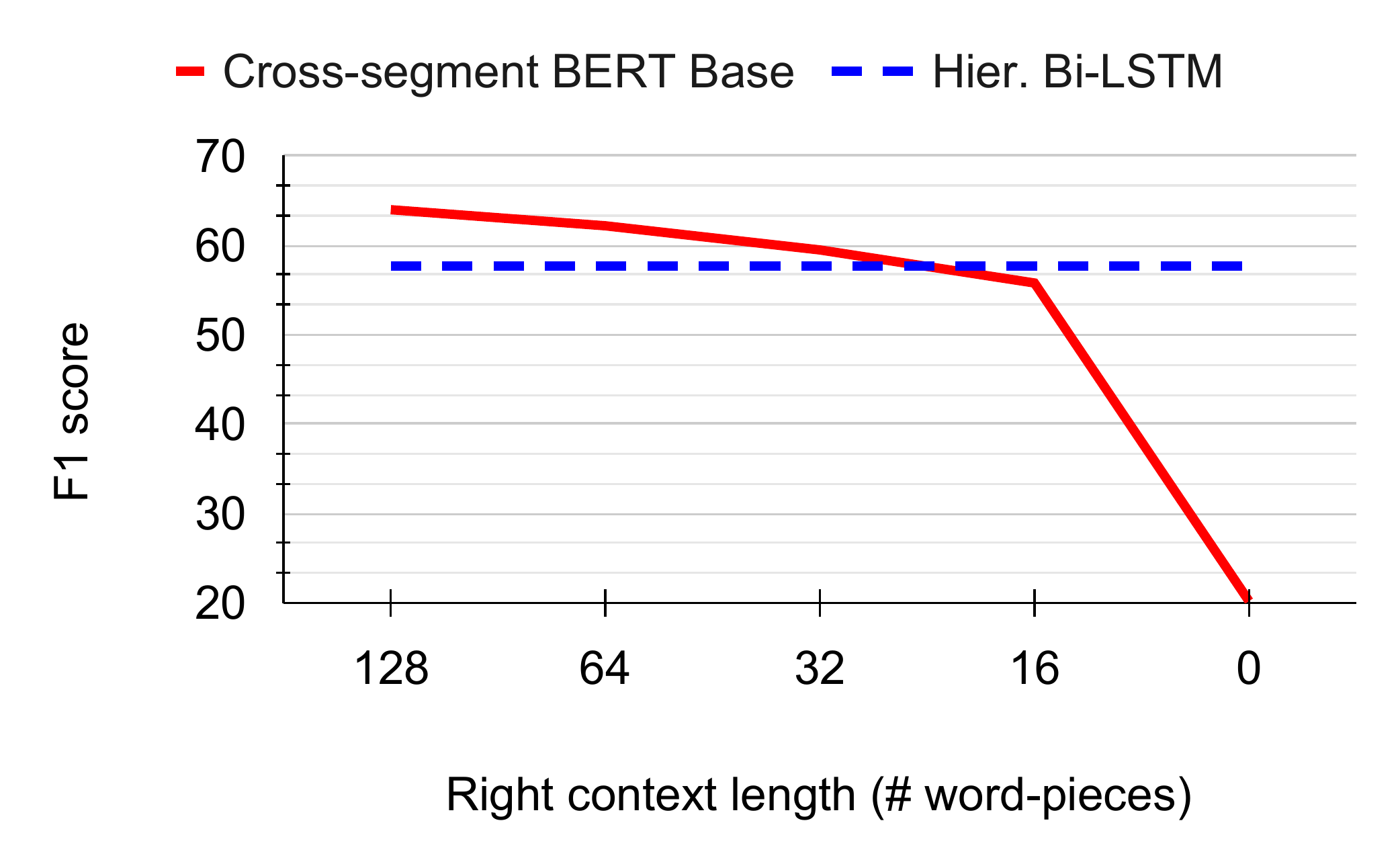}
\caption{Analysis of the importance of the right context length (solid red line). Dashed blue line denotes the hierarchical Bi-LSTM baseline encoding the full context \cite{koshorek18}.}
\label{fig:trailingContext}
\end{figure}

For the aforementioned application, it would be helpful to use as little trailing (after-the-break) context as possible. This way, we can suggest section breaks sooner. Reducing the context size also speeds up the model (as cost is quadratic in sequence length). To this end, we study the effect of trailing context size, going from 128 word-piece tokens down to 0. For this set of experiments, we held the leading context size fixed at 128 tokens, and tuned BERT$_\textrm{\sc Base}$ with a batch size of 1536 examples and a learning rate of 5e-5. The results for these 128-n experiments are shown in \Cref{fig:trailingContext}.


While the results are intuitive, it is not clear whether the performance drops because of smaller trailing context or because of smaller overall context. To answer this, we ran another experiment with 256 tokens on the left and 0 tokens on the right (256-0). 
With all else being the same, this 256-0 experiment attains F1 score of 20.2. This is much smaller than 64.0 F1 with 128 tokens on each side of the proposed break. Clearly, it is crucial that the model sees both sides of the break. This aligns with the intuition that word distributions before and after a true segment break are typically quite different \cite{hearst97}. However, presenting the model with just the distributions of tokens on either side of the proposed break leads to poor performance: in another experiment, we replaced the running text on either side with a sorted list of 128 most frequent tokens seen in a larger context (256 tokens) on either side, padding as necessary, and tuned BERT$_\textrm{\sc Base}$ with all else the same. This 128-128 experiment attains 39.1 F1 score, compared to 64.0 with 128-128 running text on either side. This suggests that high-performing models are doing more than just counting tokens on each side to detect semantic shift.


\subsection{Role of Transformer architecture}

\begin{table}
\footnotesize
\begin{center}
\begin{tabular}{lrr}
\toprule 
Architecture & Parameters & F1 \\
\midrule 
L24-H1024-A16 & 336M & 66.0 \\
L12-H768-A12 & 110M & 64.0 \\
L12-H512-A8 & 54M & 63.4 \\
L12-H256-A8 & 17M & 62.3 \\
L6-H256-A8 & 13M & 60.2 \\
L4-H256-A4 & 11M & 58.2 \\
L12-H128-A8 & 6M & 59.2 \\
L6-H128-A8 & 5M & 57.9 \\
L12-H64-A8 & 2.6M & 55.5 \\
\bottomrule 
\end{tabular}
\end{center}
\caption{Effect of model architecture on Wiki-727K results.} 
\label{tab:results-wiki-arch}
\end{table}

The best cross-segment BERT model relies on BERT$_\textrm{Large}$. While powerful, this model is slow and expensive to run. For large-scale applications such as offline analysis for web search or online document processing such as Google Docs or Microsoft Office, such large models are prohibitively expensive.  \Cref{tab:results-wiki-arch} shows the effect of model size on performance. For these experiments, we initialized the training with models pre-trained as in the BERT paper \citep{BertPaper}. The first two experiments are initialized with BERT$_\textrm{\sc Large}$ and BERT$_\textrm{\sc Base}$ respectively.

Overall, the larger the model, the better the performance. These experiments also suggest that, in addition to the size, the configuration also matters. A 128-dimensional model with more layers can outperform a 256-dimensional model with fewer layers. While the new state-of-the-art is several standard deviations better than the previous one (as reported in \Cref{tab:results-all}), this gain came at a steep cost in the model size. This is unsatisfactory, as large size hinders the possibility of using the model at scale and with low latency, which is desirable for this application \cite{wang:emnlp18}. 
In the next section, we explore smaller models with better performance using model distillation.


\subsection{Model distillation}

As can be seen from the previous section, performance degrades quite quickly as smaller and therefore more practical networks are used. An alternative to the pre-training/fine-tuning approach used above is distillation, which is a popular technique to build small networks \cite{bucila06,hinton15}. Instead of training directly a small model on the segmentation data with binary labels, we can instead leverage the knowledge learnt by our best network ---called in this context the 'teacher'--- as follows. First, we record the predictions, or more precisely the output logits, from the teacher model on the full dataset. Then, a small 'student' model is trained using a combination of a cross-entropy loss with the true labels, and a MSE loss to mimick the teacher logits. The relative weight between the two objectives is treated as a hyperparameter.


\begin{table}
\footnotesize
\begin{center}
\begin{tabular}{lrr}
\toprule 
Architecture & Parameters & F1 \\
\midrule 
L4-H256-A4 & 11M & 63.0 \\
L6-H128-A8 & 5M & 62.5 \\
\bottomrule 
\end{tabular}
\end{center}
\caption{Distillation results on the Wiki-727K dataset.}
\label{tab:results-wiki-distill}
\end{table}


Distillation results are presented in \Cref{tab:results-wiki-distill}.
We can see that the distilled models perform better than models trained directly on the training data without a teacher, increasing F1-scores by over 4 points.
We notice that distillation allows much more compact models to significantly outperform the previous state-of-the-art.
Unfortunately, we cannot directly compare model sizes with \cite{koshorek18}
since they rely on a subset of the embeddings from
a public word2vec archive that includes over 3M vocabulary items,
including phrases, most of which are likely never used by the model. It is however fair to say their 
hierarchical Bi-LSTM model relies on dozens of millions
of embedding parameters (even though these are not fine-tuned during training) as well as several
million LSTM parameters.

\section{Conclusion}

In this paper, we introduce three new model architectures
for text segmentation tasks: a cross-segment BERT model
that uses only local context around candidate breaks,
as well as two hierarchical models, BERT+Bi-LSTM
and hierarchical BERT.
We evaluated these three models on document and discourse 
segmentation using three standard datasets, and 
compared them with other recent neural approaches.
Our experiments showed that all of our models improve 
the current state-of-the-art.
In particular, we found that a cross-segment BERT model 
is extremely competitive with hierarchical models 
which have been the focus of recent research efforts \cite{ChalkidisAA19,zhang19}.
This is surprising as it suggests that local context is
sufficient in many cases. Due to its simplicity, 
we suggest at least trying it 
as a baseline when tackling other segmentation problems
and datasets.

Naturally these results do not imply that hierarchical models
should be disregarded. We showed they are strong contenders
and we are convinced there are applications
where local context is not sufficient. 
We tried several encoders at the upper-level of the 
hierarchy.
Our experiments suggest that deep transformer encoders
are useful for encoding long and complex inputs, e.g.,
documents for document segmentation applications, while Bi-LSTMs proved useful for discourse segmentation.
Moreover, RNNs in general may also be useful for very long
documents as they are able to deal with very long input sequences.

Finally, we performed ablation studies to better understand
the role of context and model size. 
Consequently, we showed that distillation
is an effective technique to build much more compact models
to use in practical settings.

In future work, we plan to further investigate how different techniques apply to the problem of text segmentation, including data augmentation \cite{wei-zou-2019-eda,seq2sesls} and methods for regularization and mitigating labeling noise \cite{jiang2020smart,lukasik2020does}.

\bibliography{bibliography}

\begin{thebibliography}{37}
\expandafter\ifx\csname natexlab\endcsname\relax\def\natexlab#1{#1}\fi

\bibitem[{Badjatiya et~al.(2018)Badjatiya, Kurisinkel, Gupta, and
  Varma}]{badjatiya2018}
Pinkesh Badjatiya, Litton~J. Kurisinkel, Manish Gupta, and Vasudeva Varma.
  2018.
\newblock \href {http://arxiv.org/abs/1808.09935} {Attention-based neural text
  segmentation}.
\newblock \emph{CoRR}, abs/1808.09935.

\bibitem[{Beeferman et~al.(1999)Beeferman, Berger, and Lafferty}]{beeferman99}
Doug Beeferman, Adam Berger, and John Lafferty. 1999.
\newblock Statistical models for text segmentation.
\newblock \emph{Machine Learning}, 34(1):177--210.

\bibitem[{Bucila et~al.(2006)Bucila, Caruana, and Niculescu{-}Mizil}]{bucila06}
Cristian Bucila, Rich Caruana, and Alexandru Niculescu{-}Mizil. 2006.
\newblock \href {https://doi.org/10.1145/1150402.1150464} {Model compression}.
\newblock In \emph{Proceedings of the Twelfth {ACM} {SIGKDD} International
  Conference on Knowledge Discovery and Data Mining, Philadelphia, PA, USA,
  August 20-23, 2006}, pages 535--541.

\bibitem[{Carlson et~al.(2001)Carlson, Marcu, and
  Okurovsky}]{carlson:sigdial2001}
Lynn Carlson, Daniel Marcu, and Mary~Ellen Okurovsky. 2001.
\newblock \href {https://www.aclweb.org/anthology/W01-1605} {Building a
  discourse-tagged corpus in the framework of rhetorical structure theory}.
\newblock In \emph{Proceedings of the Second {SIG}dial Workshop on Discourse
  and Dialogue}.

\bibitem[{Chalkidis et~al.(2019)Chalkidis, Androutsopoulos, and
  Aletras}]{ChalkidisAA19}
Ilias Chalkidis, Ion Androutsopoulos, and Nikolaos Aletras. 2019.
\newblock \href {https://www.aclweb.org/anthology/P19-1424/} {Neural legal
  judgment prediction in english}.
\newblock In \emph{Proceedings of the 57th Conference of the Association for
  Computational Linguistics, {ACL} 2019, Florence, Italy, July 28- August 2,
  2019, Volume 1: Long Papers}, pages 4317--4323.

\bibitem[{Choi(2000)}]{Choi2000}
Freddy Y.~Y. Choi. 2000.
\newblock \href {http://dl.acm.org/citation.cfm?id=974305.974309} {Advances in
  domain independent linear text segmentation}.
\newblock In \emph{Proceedings of the 1st North American Chapter of the
  Association for Computational Linguistics Conference}, NAACL 2000, pages
  26--33, Stroudsburg, PA, USA. Association for Computational Linguistics.

\bibitem[{Devlin et~al.(2018)Devlin, Chang, Lee, and Toutanova}]{BertPaper}
Jacob Devlin, Ming{-}Wei Chang, Kenton Lee, and Kristina Toutanova. 2018.
\newblock \href {http://arxiv.org/abs/1810.04805} {{BERT:} pre-training of deep
  bidirectional transformers for language understanding}.
\newblock \emph{CoRR}, abs/1810.04805.

\bibitem[{Eisenstein(2009)}]{eisenstein09}
Jacob Eisenstein. 2009.
\newblock Hierarchical text segmentation from multi-scale lexical cohesion.
\newblock In \emph{Human Language Technologies: Conference of the North
  American Chapter of the Association of Computational Linguistics,
  Proceedings,}, pages 353--361.

\bibitem[{Hearst(1997)}]{hearst97}
Marti Hearst. 1997.
\newblock {TextTiling: segmenting text into multi-paragraph subtopic passages}.
\newblock \emph{Computational Linguistics}, 23(1):33--64.

\bibitem[{Hinton et~al.(2015)Hinton, Vinyals, and Dean}]{hinton15}
Geoffrey Hinton, Oriol Vinyals, and Jeffrey Dean. 2015.
\newblock \href {http://arxiv.org/abs/1503.02531} {Distilling the knowledge in
  a neural network}.
\newblock In \emph{NIPS Deep Learning and Representation Learning Workshop}.

\bibitem[{Jiang et~al.(2020)Jiang, He, Chen, Liu, Gao, and
  Zhao}]{jiang2020smart}
Haoming Jiang, Pengcheng He, Weizhu Chen, Xiaodong Liu, Jianfeng Gao, and Tuo
  Zhao. 2020.
\newblock \href {http://arxiv.org/abs/1911.03437} {Smart: Robust and efficient
  fine-tuning for pre-trained natural language models through principled
  regularized optimization}.

\bibitem[{Joty et~al.(2019)Joty, Carenini, Ng, and
  Murray}]{joty-etal-2019-discourse}
Shafiq Joty, Giuseppe Carenini, Raymond Ng, and Gabriel Murray. 2019.
\newblock \href {https://doi.org/10.18653/v1/P19-4003} {Discourse analysis and
  its applications}.
\newblock In \emph{Proceedings of the 57th Annual Meeting of the Association
  for Computational Linguistics: Tutorial Abstracts}, pages 12--17, Florence,
  Italy. Association for Computational Linguistics.

\bibitem[{Joty et~al.(2015)Joty, Carenini, and Ng}]{joty-etal-2015-codra}
Shafiq Joty, Giuseppe Carenini, and Raymond~T. Ng. 2015.
\newblock \href {https://doi.org/10.1162/COLI_a_00226} {{CODRA}: A novel
  discriminative framework for rhetorical analysis}.
\newblock \emph{Computational Linguistics}, 41(3):385--435.

\bibitem[{Koshorek et~al.(2018)Koshorek, Cohen, Mor, Rotman, and
  Berant}]{koshorek18}
Omri Koshorek, Adir Cohen, Noam Mor, Michael Rotman, and Jonathan Berant. 2018.
\newblock \href {http://arxiv.org/abs/1803.09337} {Text segmentation as a
  supervised learning task}.
\newblock \emph{CoRR}, abs/1803.09337.

\bibitem[{Li et~al.(2018)Li, Sun, and Joty}]{segbot}
Jing Li, Aixin Sun, and Shafiq Joty. 2018.
\newblock \href {https://doi.org/10.24963/ijcai.2018/579} {Segbot: A generic
  neural text segmentation model with pointer network}.
\newblock In \emph{Proceedings of the Twenty-Seventh International Joint
  Conference on Artificial Intelligence, {IJCAI-18}}, pages 4166--4172.
  International Joint Conferences on Artificial Intelligence Organization.

\bibitem[{Li et~al.(2016)Li, Thadani, and Stent}]{li-etal-2016-role}
Junyi~Jessy Li, Kapil Thadani, and Amanda Stent. 2016.
\newblock \href {https://doi.org/10.18653/v1/W16-3617} {The role of discourse
  units in near-extractive summarization}.
\newblock In \emph{Proceedings of the 17th Annual Meeting of the Special
  Interest Group on Discourse and Dialogue}, pages 137--147, Los Angeles.
  Association for Computational Linguistics.

\bibitem[{Llopis et~al.(2002)Llopis, Rodr\'{\i}guez, and
  Gonz\'{a}lez}]{Llopis2002}
Fernando Llopis, Antonio~Ferr\'{a}ndez Rodr\'{\i}guez, and Jos{\'e} Luis~Vicedo
  Gonz\'{a}lez. 2002.
\newblock \href {http://dl.acm.org/citation.cfm?id=647344.724145} {Text
  segmentation for efficient information retrieval}.
\newblock In \emph{Proceedings of the Third International Conference on
  Computational Linguistics and Intelligent Text Processing}, CICLing '02,
  pages 373--380, Berlin, Heidelberg. Springer-Verlag.

\bibitem[{Loshchilov and Hutter(2017)}]{adamw18}
Ilya Loshchilov and Frank Hutter. 2017.
\newblock \href {http://arxiv.org/abs/1711.05101} {Fixing weight decay
  regularization in adam}.
\newblock \emph{CoRR}, abs/1711.05101.

\bibitem[{Lukasik et~al.(2020{\natexlab{a}})Lukasik, Bhojanapalli, Menon, and
  Kumar}]{lukasik2020does}
Michal Lukasik, Srinadh Bhojanapalli, Aditya~Krishna Menon, and Sanjiv Kumar.
  2020{\natexlab{a}}.
\newblock Does label smoothing mitigate label noise?
\newblock \emph{arXiv preprint arXiv:2003.02819}.

\bibitem[{Lukasik et~al.(2020{\natexlab{b}})Lukasik, Jain, Menon, Kim,
  Bhojanapalli, Yu, and Kumar}]{seq2sesls}
Michal Lukasik, Himanshu Jain, Aditya Menon, Seungyeon Kim, Srinadh
  Bhojanapalli, Felix Yu, and Sanjiv Kumar. 2020{\natexlab{b}}.
\newblock Semantic label smoothing for sequence to sequence problems.
\newblock In \emph{Proceedings of the 2020 Conference on Empirical Methods in
  Natural Language Processing}.

\bibitem[{Lukasik and Zens(2018)}]{lukasik2018}
Michal Lukasik and Richard Zens. 2018.
\newblock \href {https://doi.org/10.18653/v1/D18-1374} {Content explorer:
  Recommending novel entities for a document writer}.
\newblock In \emph{Proceedings of the 2018 Conference on Empirical Methods in
  Natural Language Processing}, pages 3371--3380, Brussels, Belgium.
  Association for Computational Linguistics.

\bibitem[{Mann and Thompson(1988)}]{mann1988}
William~C Mann and Sandra~A Thompson. 1988.
\newblock Rhetorical structure theory: Toward a functional theory of text
  organization.
\newblock \emph{Text - Interdisciplinary Journal for the Study of Discourse},
  8(3):243--281.

\bibitem[{Marcus et~al.(1994)Marcus, Kim, Marcinkiewicz, MacIntyre, Bies,
  Ferguson, Katz, and Schasberger}]{Marcus:1994:PTA:1075812.1075835}
Mitchell Marcus, Grace Kim, Mary~Ann Marcinkiewicz, Robert MacIntyre, Ann Bies,
  Mark Ferguson, Karen Katz, and Britta Schasberger. 1994.
\newblock \href {https://doi.org/10.3115/1075812.1075835} {The penn treebank:
  Annotating predicate argument structure}.
\newblock In \emph{Proceedings of the Workshop on Human Language Technology},
  HLT '94, pages 114--119, Stroudsburg, PA, USA. Association for Computational
  Linguistics.

\bibitem[{Mota et~al.(2019)Mota, Eskenazi, and Coheur}]{mota19}
Pedro Mota, Maxine Eskenazi, and Lu{\'\i}sa Coheur. 2019.
\newblock {B}eam{S}eg: A joint model for multi-document segmentation and topic
  identification.
\newblock In \emph{Proceedings of the 23rd Conference on Computational Natural
  Language Learning (CoNLL)}, pages 582--592. Association for Computational
  Linguistics.

\bibitem[{Pennington et~al.(2014)Pennington, Socher, and
  Manning}]{pennington-etal-2014-glove}
Jeffrey Pennington, Richard Socher, and Christopher Manning. 2014.
\newblock \href {https://doi.org/10.3115/v1/D14-1162} {{G}love: Global vectors
  for word representation}.
\newblock In \emph{Proceedings of the 2014 Conference on Empirical Methods in
  Natural Language Processing ({EMNLP})}, pages 1532--1543, Doha, Qatar.
  Association for Computational Linguistics.

\bibitem[{Peters et~al.(2018)Peters, Neumann, Iyyer, Gardner, Clark, Lee, and
  Zettlemoyer}]{elmopaper}
Matthew~E. Peters, Mark Neumann, Mohit Iyyer, Matt Gardner, Christopher Clark,
  Kenton Lee, and Luke Zettlemoyer. 2018.
\newblock \href {http://arxiv.org/abs/1802.05365} {Deep contextualized word
  representations}.
\newblock \emph{CoRR}, abs/1802.05365.

\bibitem[{Pevzner and Hearst(2002)}]{pevzner02}
Lev Pevzner and Marti~A. Hearst. 2002.
\newblock A critique and improvement of an evaluation metric for text
  segmentation.
\newblock \emph{Comput. Linguist.}, 28(1):19–36.

\bibitem[{Shtekh et~al.(2018)Shtekh, Kazakova, Nikitinsky, and
  Skachkov}]{Shtekh2018}
Gennady Shtekh, Polina Kazakova, Nikita Nikitinsky, and Nikolay Skachkov. 2018.
\newblock \href {https://doi.org/10.1145/3290621.3290630} {Applying topic
  segmentation to document-level information retrieval}.
\newblock In \emph{Proceedings of the 14th Central and Eastern European
  Software Engineering Conference Russia}, CEE-SECR '18, pages 6:1--6:6, New
  York, NY, USA. ACM.

\bibitem[{Soricut and Marcu(2003)}]{soricut-marcu-2003-sentence}
Radu Soricut and Daniel Marcu. 2003.
\newblock \href {https://www.aclweb.org/anthology/N03-1030} {Sentence level
  discourse parsing using syntactic and lexical information}.
\newblock In \emph{Proceedings of the 2003 Human Language Technology Conference
  of the North {A}merican Chapter of the Association for Computational
  Linguistics}, pages 228--235.

\bibitem[{Utiyama and Isahara(2001)}]{utiyama01}
Masao Utiyama and Hitoshi Isahara. 2001.
\newblock A statistical model for domain-independent text segmentation.
\newblock In \emph{Proceedings of the 39th Annual Meeting on Association for
  Computational Linguistics}, pages 499--506.

\bibitem[{Vaswani et~al.(2017)Vaswani, Shazeer, Parmar, Uszkoreit, Jones,
  Gomez, Kaiser, and Polosukhin}]{Attention2017}
Ashish Vaswani, Noam Shazeer, Niki Parmar, Jakob Uszkoreit, Llion Jones,
  Aidan~N Gomez, \L~ukasz Kaiser, and Illia Polosukhin. 2017.
\newblock \href
  {http://papers.nips.cc/paper/7181-attention-is-all-you-need.pdf} {Attention
  is all you need}.
\newblock In I.~Guyon, U.~V. Luxburg, S.~Bengio, H.~Wallach, R.~Fergus,
  S.~Vishwanathan, and R.~Garnett, editors, \emph{Advances in Neural
  Information Processing Systems 30}, pages 5998--6008. Curran Associates, Inc.

\bibitem[{Vinyals et~al.(2015)Vinyals, Fortunato, and Jaitly}]{Vinyals2015}
Oriol Vinyals, Meire Fortunato, and Navdeep Jaitly. 2015.
\newblock \href {http://dl.acm.org/citation.cfm?id=2969442.2969540} {Pointer
  networks}.
\newblock In \emph{Proceedings of the 28th International Conference on Neural
  Information Processing Systems - Volume 2}, NIPS'15, pages 2692--2700,
  Cambridge, MA, USA. MIT Press.

\bibitem[{Wang et~al.(2018)Wang, Li, and Yang}]{wang:emnlp18}
Yizhong Wang, Sujian Li, and Jingfeng Yang. 2018.
\newblock Toward fast and accurate neural discourse segmentation.
\newblock In \emph{Proceedings of the 2018 Conference on Empirical Methods in
  Natural Language Processing {EMNLP-18}}, pages 962--967. Association for
  Computational Linguistics.

\bibitem[{Wei and Zou(2019)}]{wei-zou-2019-eda}
Jason Wei and Kai Zou. 2019.
\newblock \href {https://doi.org/10.18653/v1/D19-1670} {{EDA}: Easy data
  augmentation techniques for boosting performance on text classification
  tasks}.
\newblock In \emph{Proceedings of the 2019 Conference on Empirical Methods in
  Natural Language Processing and the 9th International Joint Conference on
  Natural Language Processing (EMNLP-IJCNLP)}, pages 6382--6388, Hong Kong,
  China. Association for Computational Linguistics.

\bibitem[{Wu et~al.(2016)Wu, Schuster, Chen, Le, Norouzi, Macherey, Krikun,
  Cao, Gao, Macherey, Klingner, Shah, Johnson, Liu, Łukasz Kaiser, Gouws,
  Kato, Kudo, Kazawa, Stevens, Kurian, Patil, Wang, Young, Smith, Riesa,
  Rudnick, Vinyals, Corrado, Hughes, and Dean}]{wu16}
Yonghui Wu, Mike Schuster, Zhifeng Chen, Quoc~V. Le, Mohammad Norouzi, Wolfgang
  Macherey, Maxim Krikun, Yuan Cao, Qin Gao, Klaus Macherey, Jeff Klingner,
  Apurva Shah, Melvin Johnson, Xiaobing Liu, Łukasz Kaiser, Stephan Gouws,
  Yoshikiyo Kato, Taku Kudo, Hideto Kazawa, Keith Stevens, George Kurian,
  Nishant Patil, Wei Wang, Cliff Young, Jason Smith, Jason Riesa, Alex Rudnick,
  Oriol Vinyals, Greg Corrado, Macduff Hughes, and Jeffrey Dean. 2016.
\newblock \href {http://arxiv.org/abs/1609.08144} {Google's neural machine
  translation system: Bridging the gap between human and machine translation}.
\newblock \emph{CoRR}, abs/1609.08144.

\bibitem[{Xuan~Bach et~al.(2012)Xuan~Bach, Le~Minh, and
  Shimazu}]{xuan-bach-etal-2012-reranking}
Ngo Xuan~Bach, Nguyen Le~Minh, and Akira Shimazu. 2012.
\newblock \href {https://www.aclweb.org/anthology/W12-1623} {A reranking model
  for discourse segmentation using subtree features}.
\newblock In \emph{Proceedings of the 13th Annual Meeting of the Special
  Interest Group on Discourse and Dialogue}, pages 160--168, Seoul, South
  Korea. Association for Computational Linguistics.

\bibitem[{Zhang et~al.(2019)Zhang, Wei, and Zhou}]{zhang19}
Xingxing Zhang, Furu Wei, and Ming Zhou. 2019.
\newblock \href {https://doi.org/10.18653/v1/P19-1499} {{HIBERT}: Document
  level pre-training of hierarchical bidirectional transformers for document
  summarization}.
\newblock In \emph{Proceedings of the 57th Annual Meeting of the Association
  for Computational Linguistics}, pages 5059--5069.

\end{thebibliography}
\bibliographystyle{acl_natbib}

\end{document}